\documentclass{article}
\usepackage{graphicx}
\graphicspath{ {figures/} } 
\usepackage{tabularx,booktabs,makecell,ragged2e}
\newcolumntype{Y}{>{\RaggedRight\arraybackslash}X} 
\usepackage{graphicx,tabularx,booktabs,ragged2e}
\newcolumntype{Y}{>{\RaggedRight\arraybackslash}X}
\usepackage{capt-of}

 \usepackage[sglblindworkshop, final]{neurips_2025}
\workshoptitle{Metric-Fair Prompting: Treating Similar Samples Similarly}

\usepackage[utf8]{inputenc} 
\usepackage[T1]{fontenc}    
\usepackage{hyperref}       
\usepackage{url}            
\usepackage{booktabs}       
\usepackage{amsfonts}       
\usepackage{nicefrac}       
\usepackage{microtype}      
\usepackage{xcolor}         
\usepackage{amsmath,amssymb,amsfonts,amsthm}
\usepackage{mathtools}
\usepackage{booktabs}
\usepackage{microtype}
\usepackage{url}
\usepackage{graphicx}
\usepackage{wrapfig}
\usepackage{multirow}
\usepackage{enumitem}
\usepackage{algorithm}
\usepackage{algpseudocode}
\usepackage{xcolor}

\newtheorem{definition}{Definition}

\title{Metric-Fair Prompting: Treating Similar Samples Similarly}

\author{
	Jing Wang \\
	NLM \\
	\texttt{jing.wang20@nih.gov}
	\and
	Jie Shen \\
	Stevens Institute of Technology\\
\texttt{jie.shen@stevens.edu}
	\and
	 Xing Niu\\
 AWS AI \\
 \texttt{xingniu@amazon.com}
 \and 
 Tong Zhang\\
 University of Illinois Urbana-Champaign \\
 \texttt{tongzhang@tongzhang-ml.org}
 \and
 Jeremy Weiss\\
 NLM\\
	\texttt{jeremy.weiss@nih.gov}
}

\begin{document}
	
	\maketitle
	
	\begin{abstract}
		We introduce \emph{Metric-Fair Prompting}, a fairness-aware prompting framework that guides large language models (LLMs) to make decisions under metric-fairness constraints. In the application of multiple-choice medical question answering, each {(question, option)} pair is treated as a binary instance with label $+1$ (correct) or $-1$ (incorrect). To promote {individual fairness}~--~treating similar instances similarly~--~we compute question similarity using NLP embeddings and solve items in \emph{joint pairs of similar questions} rather than in isolation. The prompt enforces a global decision protocol: extract decisive clinical features, map each \((\text{question}, \text{option})\) to a score $f(x)$ that acts as confidence, and impose a Lipschitz-style constraint so that similar inputs receive similar scores and, hence, consistent outputs.  Evaluated on the {MedQA (US)} benchmark, Metric-Fair Prompting is shown to improve performance over standard single-item prompting, demonstrating that fairness-guided, confidence-oriented reasoning can enhance LLM accuracy on high-stakes clinical multiple-choice questions.
		\end{abstract}
	
	\section{Introduction}
    \begin{figure}[h!]
	\centering
	\includegraphics[width=0.8\textwidth]{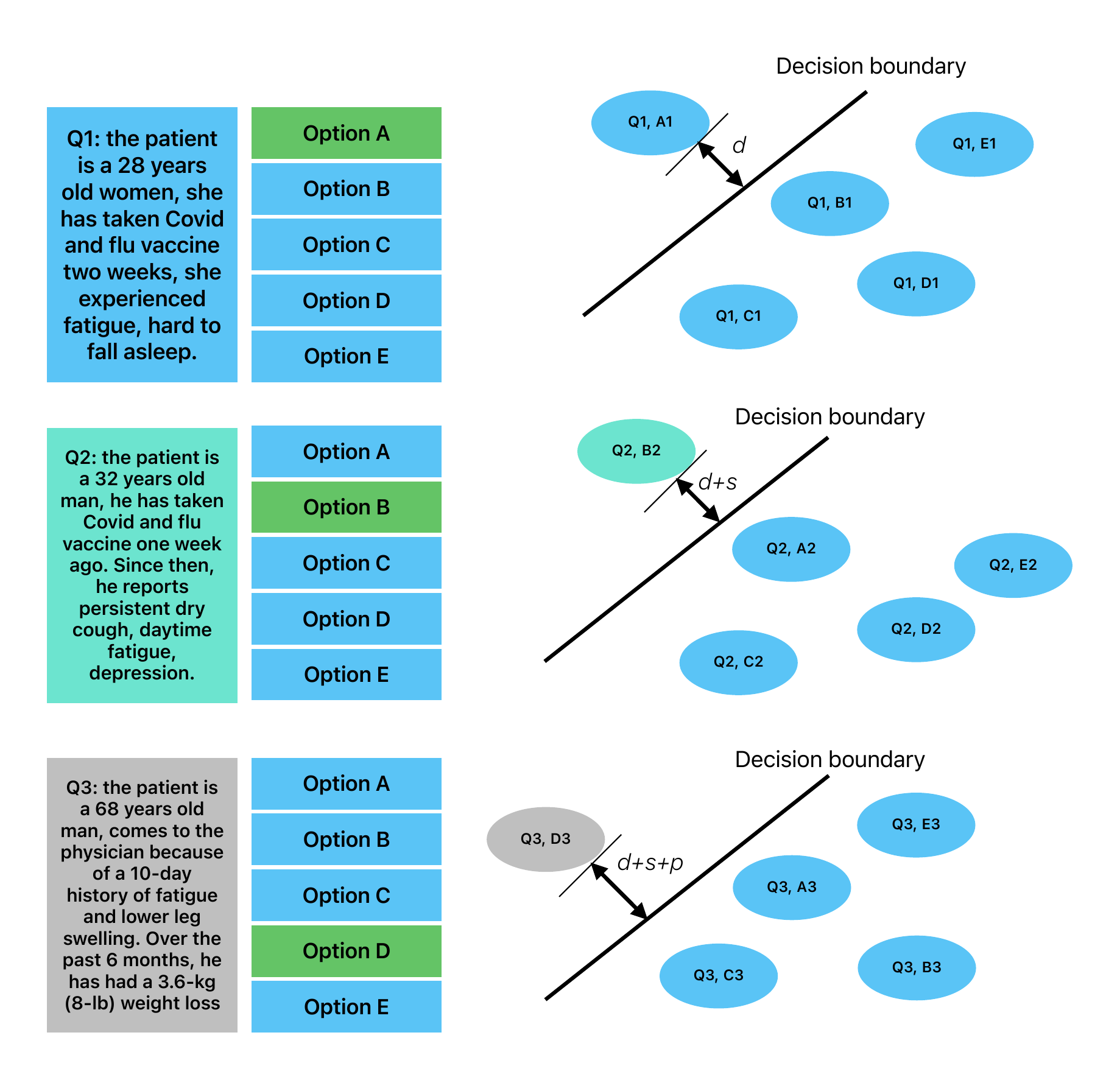}
	\caption{Geometric view of Metric-Fair Prompting. 
		Questions 1 and 2 are highly similar (small metric distance $d$); their correct options lie on the same side of the decision boundary with nearby margins $d$ and $d+s$ ($s>0$ small). 
		Question 3 is less similar to Question 1 (larger distance $d+s$, $p>0$): its correct option remains in the same half-space but with a more separated margin $d+s+p$. 
		The metric-fair (Lipschitz-like) constraint encourages similar items to receive similar scores and thus consistent decisions.}
	\label{fig:your_figure_label}
\end{figure}
Machine learning systems may disadvantage certain populations when fairness is not addressed \cite{dwork2012fairness,barocas2016big,chouldechova2018snapshot}. Such concern recently extends to large language models (LLMs), and is increasingly observed in high-stakes domains such as clinical decision making and medical examinations. In such settings, it is crucial to promote \emph{individual fairness}, i.e. treating similar instances similarly, and to base predictions on clinically determinative features rather than sensitive attributes (e.g. age, gender, race) unless those attributes are explicitly and directly relevant to the clinical task.

We study one of the most important problems: multiple-choice medical question answering (MedQA) \cite{jin2020what}, casting each \emph{(question, option)} pair as a binary classification task (\(+1\) for correct, \(-1\) for incorrect). We introduce \emph{Metric-Fair Prompting}, a prompting framework that guides an LLM under a metric-fairness constraint. Concretely, we (i) compute similarity among questions using text embeddings to identify pairs of {deterministically} similar items; (ii) present similar questions jointly so the model can enforce cross-item consistency; and (iii) ask the model to extract decisive clinical features and map each pair (question, option) to a score $f(x)$ whose magnitude can be thought of as confidence, similar to the margin in support vector machines. Technically, the metric-fairness is imposed under Lipschitz constraints on samples: if two inputs $(x, x')$ are close under a task-relevant metric \(d(x, x')\), then their scores would remain close, encouraging consistent outcomes for clinically similar cases \cite{dwork2012fairness}.

Our approach complements prior prompt-engineering methods that improve reasoning via intermediate structure, such as chain-of-thought \cite{wei2022chain}, self-consistency \cite{wang2023selfconsistency}, and search-based prompting (e.g., Tree-of-Thoughts \cite{yao2023tree} and ReAct \cite{yao2022react}). However, unlike these techniques that often typically treat items independently and optimize intra-item reasoning, Metric-Fair Prompting explicitly introduces an \emph{inter-item} coupling via a similarity metric, thereby promoting fairness through stability to small, clinically irrelevant changes.

Figure~\ref{fig:your_figure_label} shows an illustrative example. 
Questions 1 and 2 exhibit high similarity under the embedding metric (small $d$); the questions together with correct options are mapped to nearby points in a hyper-plane. 
By contrast, Question 3 is less similar to Question 1 (larger $d$); although its correct option falls on the same side of the decision boundary, it lies farther from Question 1’s point in the range space.

\paragraph{Main contributions.}
(1) We propose a fairness-aware prompting framework that treats MedQA as binary classification over \((\text{question},\text{option})\) pairs and enforces a metric-based Lipschitz constraint to encourage individual fairness. 
(2) We introduce a joint-inference protocol that feeds \emph{pairs of similar questions} to the LLM, enabling cross-item consistency and reducing near-boundary errors. 
(3) On MedQA (US), the proposed protocol improves accuracy over single-item prompting (see details in Section~\ref{sec:exp}), showing that fairness-guided, confidence-oriented reasoning can enhance LLM performance in clinical multiple-choice settings.

\begin{table}[t]
	\caption{MedQA–US examples: two patients with similar deterministic features yield similar correct options.}
	\label{tab:medqa-sim}
	\small
	\setlength{\tabcolsep}{4pt}
	\begin{tabularx}{\columnwidth}{@{}l Y@{}}
		\toprule
		\textbf{Item} & \textbf{Content} \\
		\midrule
		Question 1  (Q1)&
		A 62-year-old man presents with 5 days of fatigue, fever, and chills. He has a 9-month history of hand pain and stiffness and started a new medication 3 months ago; prior meds included ibuprofen, prednisone, and hydroxychloroquine. He does not smoke or drink. Exam: subcutaneous nodule at left elbow, old joint destruction with boutonnière deformity, no active synovitis. Labs: Hb 10.5 g/dL, WBC 3500/mm\textsuperscript{3}, platelets 100{,}000/mm\textsuperscript{3}. Which of the following is most likely to have prevented these laboratory abnormalities? \\ 
		Options for Q1 &
		A) Cobalamin \quad B) Amifostine \quad C) Pyridoxine \quad D) Leucovorin \quad E) Mesna \\ \hline \hline
		\addlinespace[0.3em]
		Question 2 (Q2) &
		A 58-year-old woman presents with 1 week of worsening fatigue and a 1-year history of hand pain and stiffness. She started a new medication 4 months ago; prior meds included ibuprofen, prednisone, and hydroxychloroquine. Exam: subcutaneous nodule at left elbow, old joint destruction with Boutonnière deformity. Labs: Hb 10.1 g/dL, WBC 3400/mm\textsuperscript{3}, platelets 101{,}000/mm\textsuperscript{3}; methylmalonic acid normal. Which of the following could have prevented these laboratory abnormalities? \\ 
		Options  for Q2&
		A) Vitamin B6 \quad B) Vitamin B12 \quad C) Amifostine \quad D) 2-Mercaptoethanesulfonate \quad E) Leucovorin \\ \hline \hline 
		\addlinespace[0.3em]
		Deterministic features &
		RA phenotype + new DMARD started months earlier + pancytopenia (anemia, leukopenia, thrombocytopenia) + normal MMA $\Rightarrow$ methotrexate-related folate pathway toxicity. Prevention: folate supplementation or folinic acid (leucovorin) rescue. \\ \hline  \hline
		Correct answers &
		Q1: D \quad Q2: E \\
		\bottomrule
	\end{tabularx}
\end{table}

\begin{figure}[t]
	\centering
	\begin{minipage}[t]{0.48\linewidth}
		\centering
		\includegraphics[width=\linewidth]{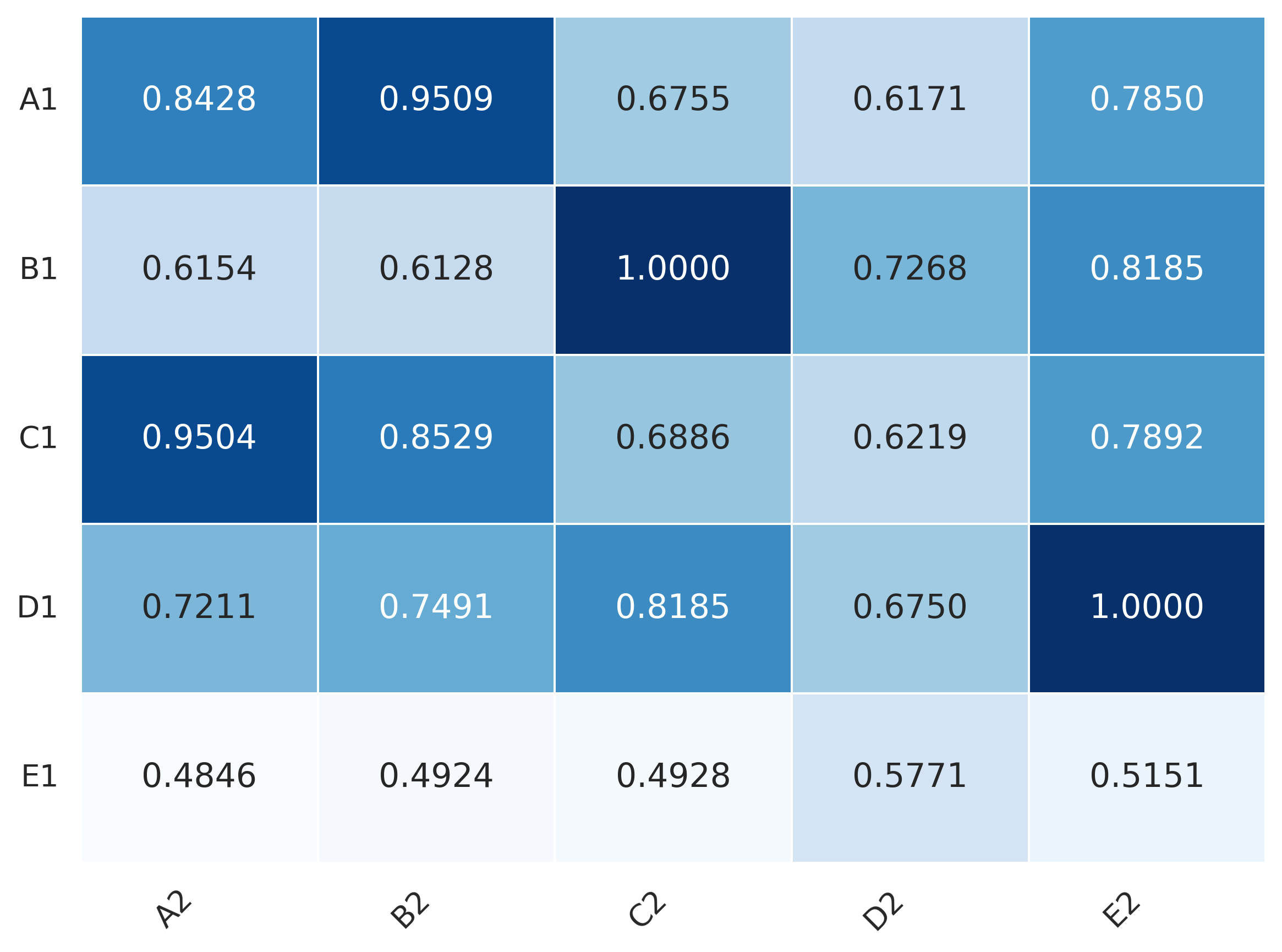}
		\caption{Correlations between the options of Question 1 and Question 2 in Table \ref{tab:medqa-sim} by Qwen3-8B embedding.}
		\label{fig:left}
	\end{minipage}\hfill
	\begin{minipage}[t]{0.48\linewidth}
		\centering
		\includegraphics[width=\linewidth]{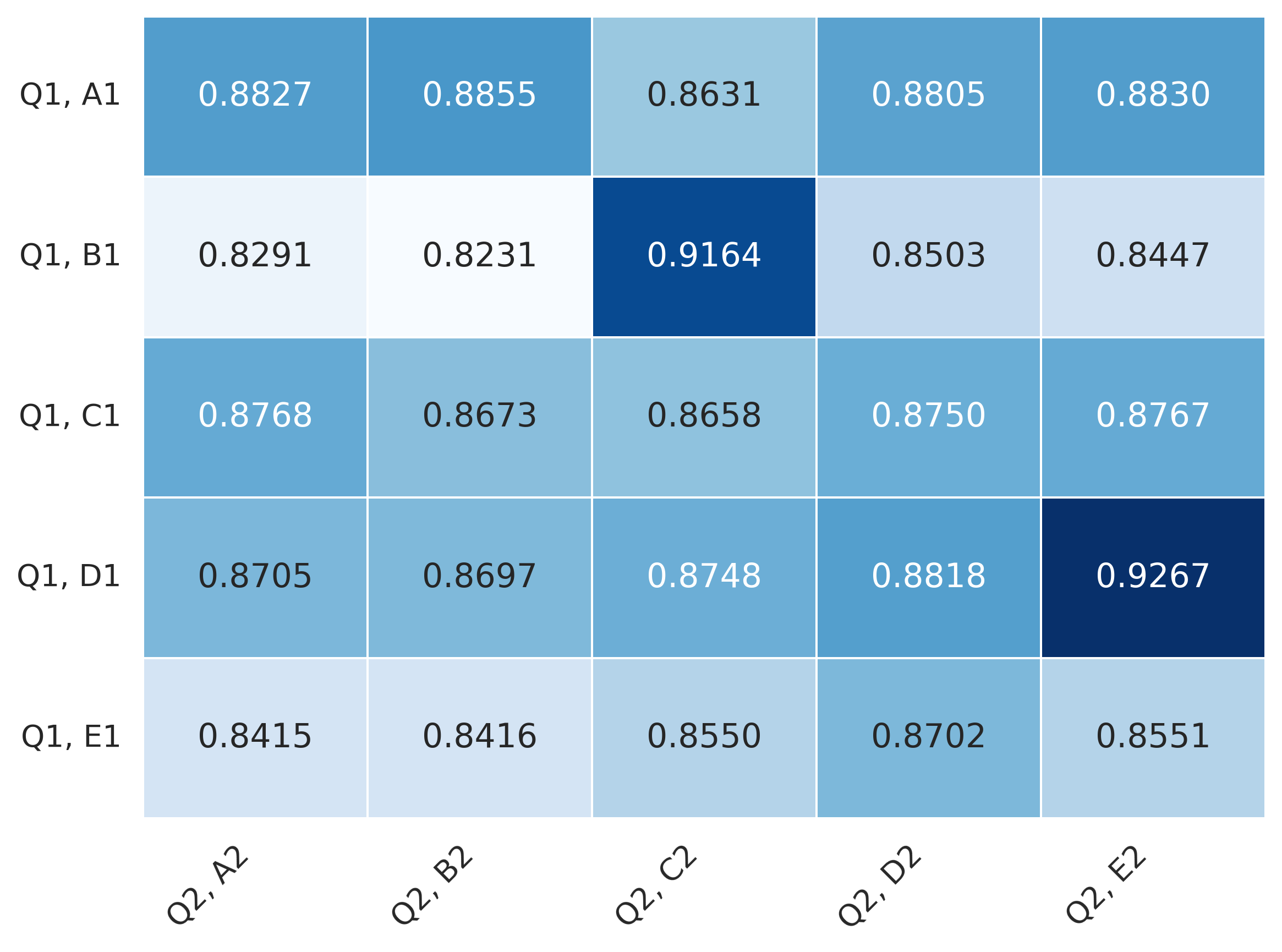}
		\caption{Correlations between questions and options from Table \ref{tab:medqa-sim} by Qwen3-8B embedding.}
		\label{fig:right}
	\end{minipage}
\end{figure}

\section{Problem Setup}

We study metric-fair learning on a domain $\mathcal{X}$ endowed with a similarity metric $d:\mathcal{X}\times\mathcal{X}\to[0,1]$. A learning algorithm receives $d$ and a set of i.i.d.\ samples from a distribution $\mathcal{D}$ over labeled examples $(x,y)\in\mathcal{X}\times\{\pm1\}$ and outputs a classifier. To accommodate fairness, we focus on {probabilistic} classifiers $h:\mathcal{X}\to[0,1]$ and interpret $h(x)$ as the probability of label $+1$ (so the probability of $-1$ equals $1-h(x)$). We refer to such probabilistic classifiers as \emph{predictors}.

We consider fairness as {treating similar individuals similarly}: two individuals that are close under $d$ should receive similar predictions. To this end, we consider learning score function $f: \mathcal{X} \to \mathbb{R}$ that is Lipschitz in the following sense.

\begin{definition}[Lipschitz mapping]
	Let $D$ be some metric on $\mathbb{R}$. A mapping $f:\mathcal{X}\to \mathbb{R}$ satisfies the $(D,d)$-Lipschitz property if for all $x, x'\in\mathcal{X}$,
	\begin{equation}
		D\big(f(x), f(x')\big)\;\le\; d(x, x').
	\end{equation}
	When $D$ and $d$ are clear from context we simply say that $f$ is {Lipschitz}.
\end{definition}

Given a loss function $L:\mathbb{R}\times\{\pm1\}\to\mathbb{R}_{\geq 0}$, our goal is to find a mapping $f$ that minimizes expected loss subject to the Lipschitz fairness constraint. This naturally leads to the optimization problem:
\begin{equation}
\min_{f}
\;\mathbb{E}_{(x,y)\sim\mathcal{D}} L(f(x), y)
\quad\text{s.t.}\quad
D\big(f(x),f(x')\big)\le d(x,x')\ \ \forall x, x' \in \mathcal{X}.
\end{equation}

For intuition we connect to linear classification. The goal is to learn a weight vector $w \in \mathbb{R}^d$ such that the prediction $\mathrm{sign}(w \cdot x)$ agrees with the label $y$ for $(x, y)$ drawn from $\mathcal{D}$. In this case, $f$ is parameterized by $w$: it maps $x$ to $w\cdot x$. For logistic regression, this score $f(x)$ will be fed to the sigmoid function to obtain the probability distribution of labels. In both cases, it is common to choose $D(\cdot, \cdot)$ as the absolute function and $d$ as certain norm.

In our application to multiple-choice medical QA, each \emph{option} paired with its \emph{question} is treated as an input $x$, and the classifier maps $(\text{question},\text{option})$ to $\{-1,1\}$ (incorrect versus correct). Metric fairness requires that if two question–option pairs are similar under $d$, then their predictive distributions (and thus their distances to the decision boundary) should also be similar; conversely, dissimilar pairs shall receive different predictions.

\section{Our Approach}

\paragraph{Motivation.}
In complex reasoning tasks such as medical examinations, distinct items can exhibit substantial semantic and clinical overlap. As illustrated in Table~\ref{tab:medqa-sim}, two stems may share \emph{deterministic} features (e.g., key signs, pathognomonic labs), and consequently their correct options tend to align \cite{wang2018provable}. 
Such prior knowledge is taken into our algorithmic design by computing similarity between questions and between {(question, option)} pairs using sentence embedding models (e.g., Qwen3-4B). 
Empirically, the correct option for a given stem has higher similarity to the stem than distractors, and stems that are similar under the embedding metric $d(\cdot,\cdot)$ often share clinically consistent answer patterns.

\paragraph{Overview.}
We propose \emph{Metric-Fair Prompting}, a joint-inference protocol that treats each \((\text{question},\text{option})\) as a binary instance and guides the LLM to behave like a margin-based classifier under a metric-fairness constraint. 
Recall that $f:\mathcal{X}\to\mathbb{R}$ is a score function. 
The predicted label is $y=\mathbf{1}\{f(x)> \alpha\}$ for some pre-determined threshold $\alpha$; for example, $\alpha = 1/2$ in logistic regression. 

\begin{table}[h]
	\caption{Metric-Fair Prompt with a Binary Margin Classifier.}
	\label{tab:two-item-fair-margin}
	\footnotesize
	\setlength{\tabcolsep}{4pt}
	\begin{tabularx}{\columnwidth}{@{}l Y@{}}
		\toprule
		\textbf{Objective} &
		Read \emph{both} questions and their options jointly; decide one option (A--E) per question.
		Output \texttt{\{"index": $i$, "answer": "A|B|C|D|E"\}} for each $i\in\{1,2\}$ (JSON only). \\
		
		\midrule
		\textbf{Formulation} &
		Consider potential correlation between the two questions. Identify deterministic clinical features that may be shared by the patients/situations. Items that are similar under a task-relevant metric should receive similar decisions (fairness-by-similarity). \\
		
		\textbf{Feature selection} &
		For each question, extract the most important features from the stem (signs, key labs/imaging, contraindications, guideline thresholds). For each option, form a feature representation $x=\phi(\text{question},\text{option})$. \\
		
		\textbf{Margin-based classifier} &
		Use a binary large-margin classifier $f:\mathcal{X}\to\mathbb{R}$ on $x$.
		The predicted label is $y=\mathbf{1}\{f(x)>0\}\in\{0,1\}$ (1 = correct), with confidence magnitude $|f(x)|$. Select, for each question, the option with the largest positive margin. \\
		
		\textbf{Fairness} &
		Let $d:\mathcal{X}\times\mathcal{X}\to[0,1]$ be a similarity metric on question–option pairs.
		Enforce a Lipschitz-like constraint so that similar inputs yield similar scores.
		If two questions are similar in decisive clinical features, prefer consistent answer patterns unless a clear clinical conflict exists. \\
		
		\textbf{Cross-item reconciliation} &
		If two options (within or across the two questions) are near the decision boundary, re-check decisive discriminators and prefer the choice that maintains cross-item consistency under $d(\cdot,\cdot)$ and standard clinical guidance. \\
		
		
		\textbf{Output format} &
		JSON only, no prose. Example:
		\[
		\texttt{[}\ \texttt{\{"index": 1, "answer": "C"\}, \{"index": 2, "answer": "A"\}}\ \texttt{]}
		\]
		\\
		
		\bottomrule
	\end{tabularx}
\end{table}

\begin{figure}[h]
	\centering
	\includegraphics[width=\linewidth]{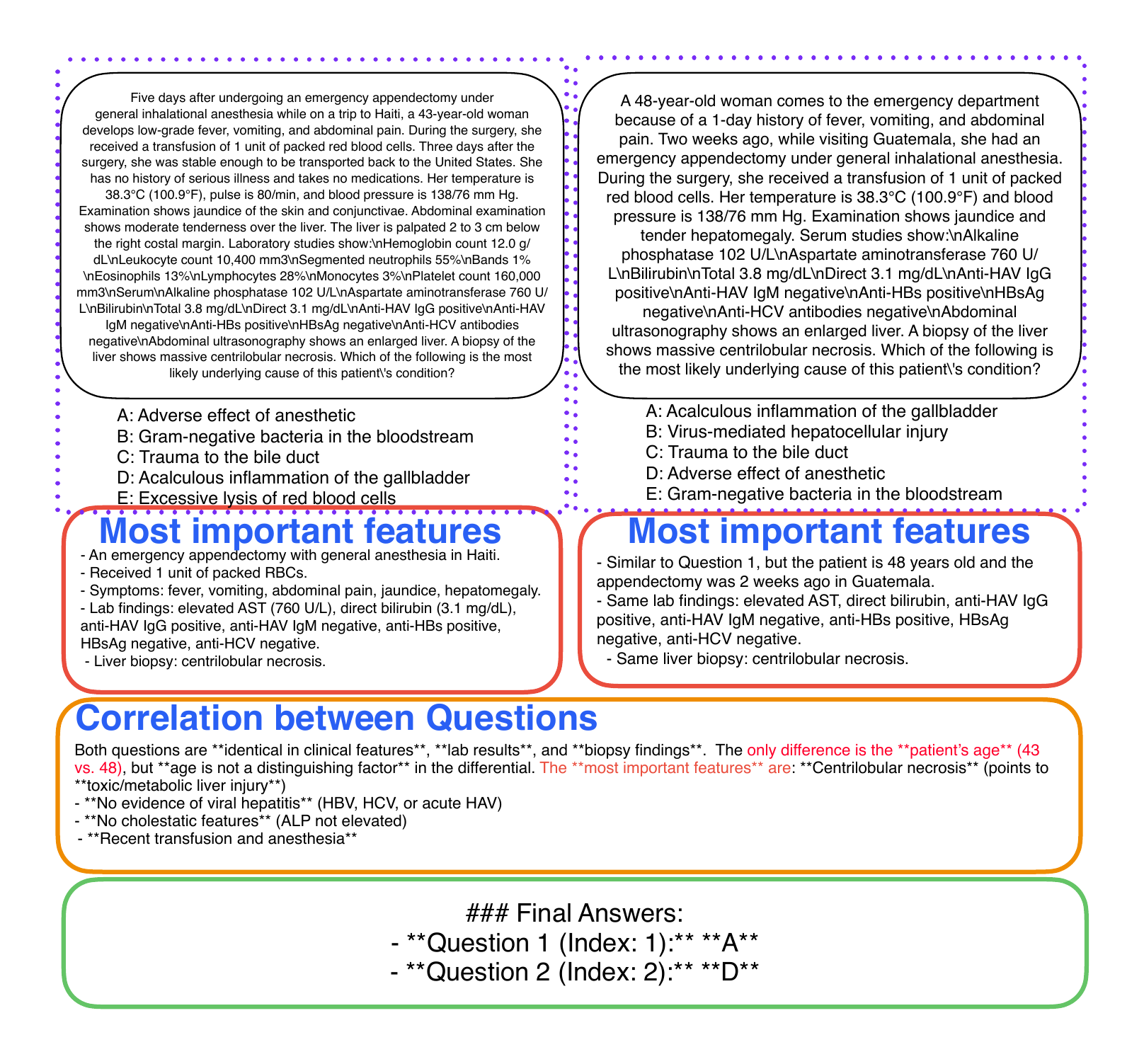}
		\vspace{-0.1in}
	\caption{Example of LLM output for two questions with cosine similarity 0. 9612 (embedding by Qwen3-4B embedding)  given our prompt. The two patients are identical in clinical features, lab results and biopsy findings. The only difference is the age, which is not a distinguishing factor. Hence the correct option is same, ``Adverse effect of anesthetic''.}
	\label{fig:example}
			\vspace{-0.1in}
\end{figure}

\begin{figure}[h]
	\centering
	\includegraphics[width=\linewidth]{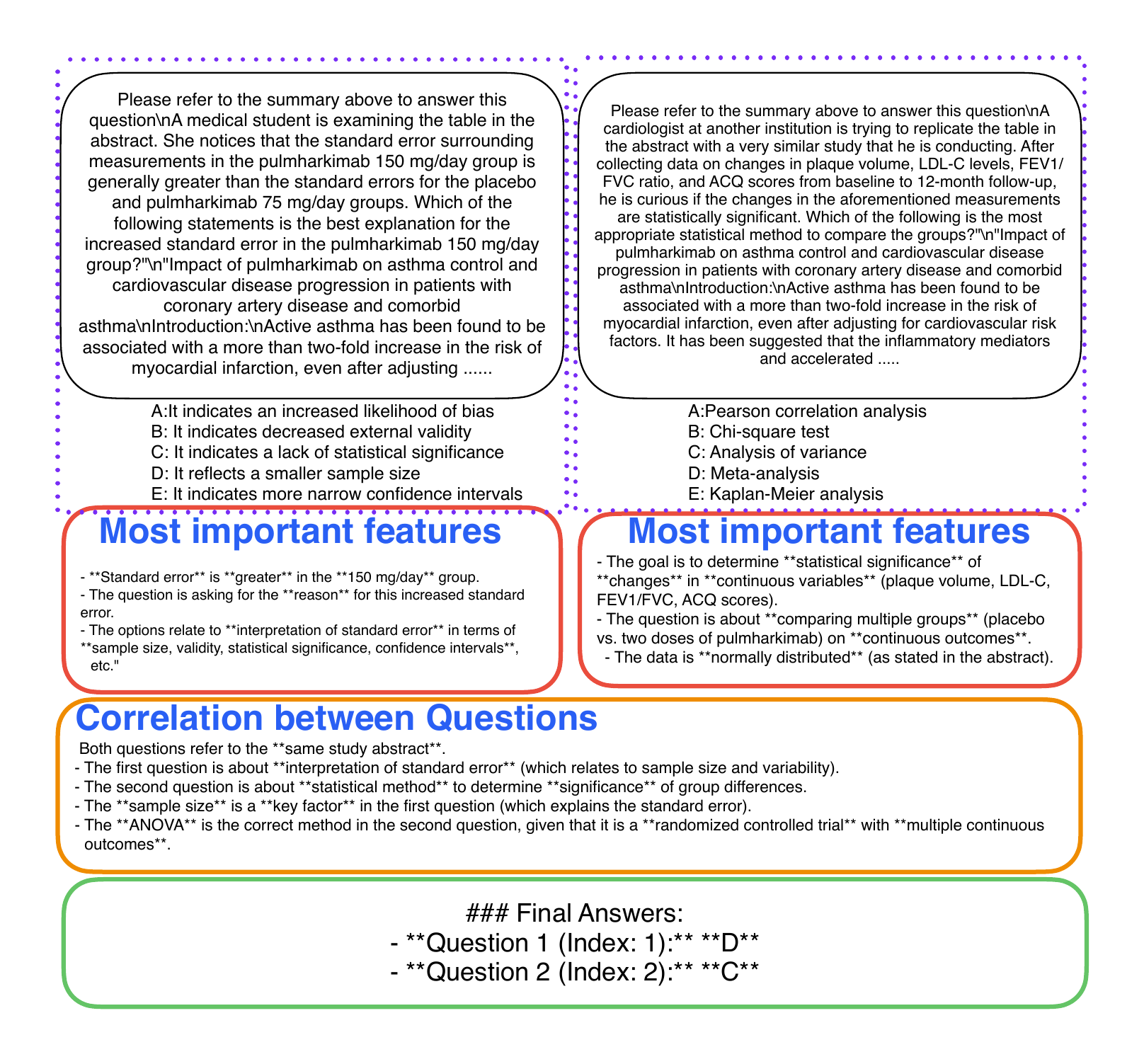}
			\vspace{-0.1in}
	\caption{Example of LLM output for two questions with cosine  similarity  0.9020 (embedding by Qwen3-4B embedding) given our prompt.  Both questions refer to the same study abstract.  The first question is about interpretation of standard error (which relates to sample size and variability). The second question is about statistical method to determine significance of group differences.}
	\label{fig:example2}
			\vspace{-0.1in}
\end{figure}

\begin{figure}[h]
	\centering
	\includegraphics[width=\linewidth]{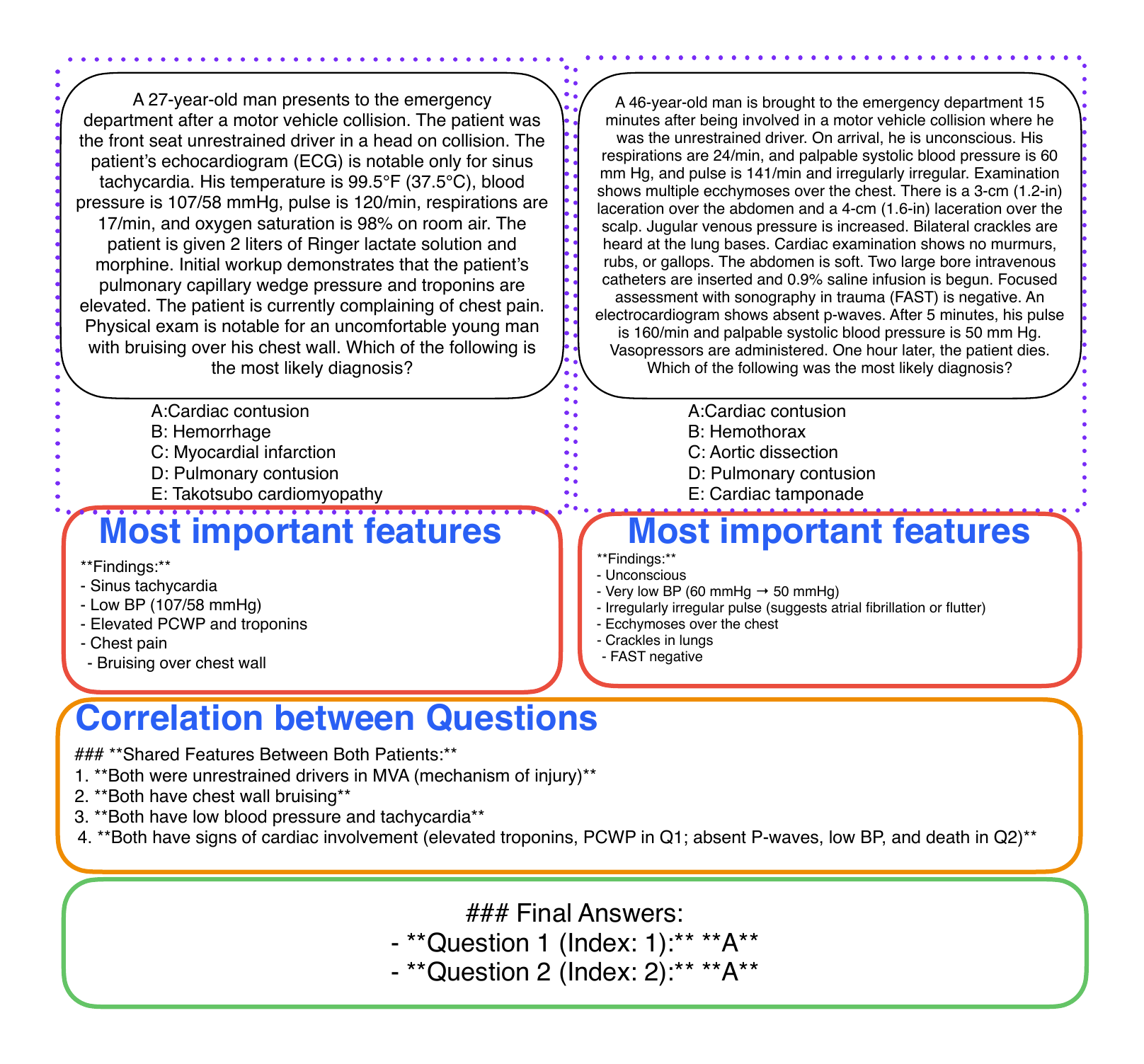}
			\vspace{-0.1in}
	\caption{Example of LLM output for two questions with cosine  similarity 0.8314 (embedding by Qwen3-4B embedding) given our prompt. Cardiac contusion (Option A) is the only option that is plausible in both cases, given similar mechanism of injury (unrestrained MVA), chest wall bruising, Elevated troponins (Q1), Cardiac dysfunction, Absence of other clear causes like MI or aortic dissection. The correct answer is same for both questions.}
	\label{fig:example3}
		\vspace{-0.1in}
\end{figure}
\paragraph{Pipeline.}
Given a set of samples, our approach consists of five major steps:
\begin{enumerate}[label=(\roman*), leftmargin=1.5em]
	\item \textbf{Pair selection.} Embed all stems; for each stem select its nearest neighbor under $d(\cdot,\cdot)$ to form a two-item batch (high-similarity pair).
	\item \textbf{Metric fairness.} Instruct the LLM that similar items should yield similar decisions (Lipschitz-like constraint), and that decisions must rely on clinically determinative features rather than sensitive attributes.
	\item \textbf{Margin/half-space reasoning.} For each \((\text{question},\text{option})\), compute the score $f(x)$. Then, eliminate clear negatives (large negative margin), and deal with near-boundary candidates using decisive clinical discriminators (guidelines, pathognomonic findings, contraindications).
	\item \textbf{Cross-item consistency.} Reconcile near ties within the pair by preferring choices that maintain consistency across similar items under $d(\cdot,\cdot)$.
	\item \textbf{Strict output.} Emit only machine-parsable results (e.g., JSON with \texttt{\{"index": $i$, "answer": "A|B|C|D|E"\}} for $i\in\{1,2\}$).
\end{enumerate}

Jointly presenting similar items introduces an inter-item coupling that stabilizes decisions near the boundary, encourages fairness via Lipschitz stability, and reduces reliance on spurious cues. 
This confidence-informed, metric-fair prompting improves robustness on clinically proximate stems while preserving a strict, parseable interface for evaluation.

\section{Experiments}
\label{sec:exp}

\paragraph{Setup.}
All experiments were conducted on a single NVIDIA RTX 6000 Ada.\footnote{No multi-GPU or model parallelism was used.} 
We evaluate Qwen models \cite{qwen-tech-report-2023} from HuggingFace (via \texttt{transformers}) with Unsloth optimizations for efficient inference.
Unless stated otherwise, we use a low temperature ($T{=}0.2$) and greedy decoding (\texttt{do\_sample{=}False}) to stabilize multiple–choice predictions.

\paragraph{Dataset.}
We use the MedQA (US) test split (\(N{=}1{,}273\) items), each with one correct option (A–E). 
We frame each \emph{(question, option)} as a binary instance (correct vs.\ incorrect).

\paragraph{Similarity and pairing.}
We embed every question with Qwen3-4B embedding and compute cosine similarity.
For each question \(q\), we select its nearest neighbor \(q'\) (excluding itself) to form a \emph{two-item} batch \((q,q')\).
This produces \(N\) pairs (some questions may appear in multiple pairs as a neighbor).
The top-3 example pairs by similarity have scores \(0.9612\), \(0.9020\), and \(0.8314\).

\paragraph{Prompting protocol.}
We apply the \emph{Metric-Fair Prompting} template in Table~\ref{tab:two-item-fair-margin}:
(i) jointly read both questions; 
(ii) enforce a metric-fair (Lipschitz-like) constraint over similar items; 
(iii) use a margin/half-space decision rule over \((\text{question},\text{option})\) features; 
(iv) reconcile near-boundary choices to maintain cross-item consistency; and 
(v) output strict JSON containing only the two answers. 

\paragraph{Conflict resolution.}
Because a question can appear in multiple pairs, it may receive two predictions.
When predictions disagree, we trigger a light-weight \emph{review prompt} that asks the model to re-evaluate both items jointly and to output (A-E) with a scalar confidence.
We keep the answer with higher confidence; if confidences tie, we prefer the answer with larger decision margin (when available) or fall back to the original single-item prediction.

\paragraph{Baselines and metric.}
The main baseline is \emph{single-item prompting} (standard instruction, one question at a time).
We report {accuracy} (\%) on the test split.

\paragraph{Results.}
Metric-Fair Prompting with Qwen3-14B improves accuracy from \(\mathbf{68\%}\) (single-item) to \(\mathbf{84\%}\) (two-item, metric-fair, joint inference), demonstrating that coupling similar items via a fairness-aware, margin-oriented protocol yields substantial gains on near-boundary decisions and promotes cross-item consistency.

\begin{table}[t]
	\centering
	\caption{MedQA (US) test accuracy (\%). Qwen3-14B with Metric-Fair Prompting substantially outperforms single-item prompting.}
	\label{tab:main-acc}
	\begin{tabular}{lcc}
		\toprule
		Model / Prompt & Single-item & Metric-Fair (two-item) \\
		\midrule
		Qwen3-14B & 68.0 & \textbf{84.0} \\
		\bottomrule
	\end{tabular}
\end{table}

\subsection{Qualitative Examples}
\label{sec:qualitative}

We illustrate how Metric-Fair Prompting enforces cross-item consistency and fairness on three highly similar question pairs selected by cosine similarity of stem embeddings (Qwen3-4B). In each case, the model reads both items jointly, extracts decisive clinical/statistical features, and applies a margin-based decision with a Lipschitz-like stability constraint.

\paragraph{Near-duplicate clinical stems (cosine = 0.9612).}
Figure~\ref{fig:example} shows two stems that are clinically indistinguishable with respect to decisive features (symptoms, exam, labs, biopsy). The only difference is age, which is \emph{not} determinative for the underlying mechanism queried. 
Under the metric-fair constraint, the model maps both (question, option) pairs to nearby points in the half-space and selects the same correct option, \emph{Adverse effect of anesthetic}. This demonstrates:
(i) \textbf{individual fairness}: similar items receive similar scores/decisions; 
(ii) \textbf{demographic robustness}: age is ignored when it is not clinically decisive; and 
(iii) \textbf{boundary stability}: small semantic perturbations do not flip the decision.

\paragraph{Shared evidence, different foci (cosine = 0.9020).}
In Figure~\ref{fig:example2}, both stems reference the same study abstract but ask distinct, yet related questions: one on interpretation of the standard error (sample size/variability), the other on choosing an inferential method for group differences. 
Joint presentation encourages the model to:
(a) build a single internal representation of the study's statistical structure; 
(b) project each item to the appropriate decision subspace (interpretation vs.\ method); and 
(c) preserve cross-item \emph{consistency} (e.g., larger $n \Rightarrow$ smaller SE; appropriate test tied to design/assumptions).
The pairwise setting reduces near-boundary errors that arise when interpreting an abstract in isolation.

\paragraph{Trauma mechanism coherence (cosine = 0.8314).}
Figure~\ref{fig:example3} presents two trauma cases with shared mechanism (unrestrained MVA) and overlapping findings (chest wall bruising; hemodynamic/ECG abnormalities). Among the options, \emph{Cardiac contusion} (A) is uniquely plausible across both stems, consistent with elevated troponins or LV dysfunction and the absence of signs pointing to MI or dissection. 

The model aligns both answers, reflecting:
(i) \textbf{mechanism-level similarity} captured by embeddings; 
(ii) \textbf{margin reconciliation} across items (selecting the option that remains viable in both contexts); and 
(iii) \textbf{fairness by similarity}: clinically equivalent trauma patterns map to consistent outcomes.

\paragraph{Methodological notes.}
For each question $q$, we retrieve its nearest neighbor $q'$ by cosine similarity on normalized sentence embeddings. The pair $(q,q')$ is fed to the LLM with the joint protocol (Table~\ref{tab:two-item-fair-margin}). The model internally clusters decisive features, eliminates clear negatives (large negative margins), and resolves near-boundary options while enforcing a Lipschitz-like stability: if $d(q,q')$ is small, then the score difference $|f(x)-f(x')|$ remains small, promoting consistent outputs unless contradicted by a decisive discriminator (e.g., a contraindication).

\paragraph{Fairness and robustness.}
Across the examples, demographic attributes (e.g., age) are used only when explicitly clinically determinative; otherwise they are down-weighted by the prompt’s fairness guard. The pairwise setting serves as a \emph{regularizer} against spurious cues: when two similar items are solved jointly, option choices that are inconsistent across the pair are penalized by the metric constraint, improving reliability on near-boundary decisions.

\section{Related Work}
There is a growing body of work attempting to study the question of algorithmic discrimination. This literature is characterized by high-level distinction, group and individual notions of fairness. We also introduce the prompt engineering related works. 

Individual fairness posits that ``similar individuals should be treated similarly'' \cite{dwork2012fairness}. This powerful guarantees is formalized via a Lipschitz condition on the classifier mapping individuals to distributions over outcomes. Recent works study study different individual level fairness in the contexts of reinforcement and online learning. study different individual level fairness in the contexts of bandit \cite{josephbetter}. \cite{shen2022metric} studies metric-fair active learning of homogeneous halfspaces \cite{shen2021sample,shen2021power}, and show that under the distribution-dependent PAC learning model. Fairness and label efficiency can be achieved simultaneously. 

Group fairness notions assume the existence of a protected attribute (e.g. gender, race), which induces a partition of the instance space into some small number of groups. A fair classifier is one that achieves parity of some statistical measure across these groups. There are prominent measures include classification rates (statistical parity \cite{feldman2015certifying}, calibration, and false positive or negative rates \cite{hardt2016equality}. \cite{woodworth2017learning} incorporating the fairness notion of  \cite{hardt2016equality} into a statistical and computational theory of learning, and proposed a relaxation of the fairness definition to make it feasible to optimize the learning objective.  

LLMs are known to have already absorbed rich commonsense that makes it possible to propose reasonable plans conditioned on problem setting \cite{diao2024active,huang2022language,wang2023plan}. The idea of Chain-of-Thought is to enrich the few-shot examples with reasoning steps \cite{wei2022chain} . There are many studies to improve the performance of DoT on complex tasks such as dynamic least-to-most prompting \cite{drozdov2022compositional}, active Prompt \cite{diao2024active}. Tree of thought approach extends planning formulations by considering multiple potentially feasible plans simultaneously at each problem-solving step \cite{yao2023tree}. \cite{madaan2023self} introduced the ``self-reflection'' mechanism, LLMs provide feedback to their generation candidates. \cite{kim2023language} introduces review steps cover the actions and states, deciding the next action.  ``self-guided decoding'' followers a tree-search procedure with leaves sampled from stochastic beam search decoding. Our prompt teaches LLMs to think like a machine learning algorithm with constraint and objective function, the constraint is the fairness metric \cite{wang2025large}. \cite{schlag2023large} embeds LLMs in an algorithmic search procedure to help solve problems like question answering step-by-step, in which relevant paragraphs that might provide answers. 
\section{Societal Impact and Limitations}

We believe that this work will have positive societal impact as it provides a new approach for using LLMs in healthcare with fairness property. One potential limitation of this work is that the performance is not completely understood from a rigorous theoretical perspective. Also, despite the observed gains, results are \emph{not always stable}. Future work will: (i) improve stability via calibrated decoding, temperature-free beam search, and ensembling; (ii) learn task-specific similarity metrics with clinical supervision; (iii) integrate confidence estimation and selective answering; (iv) study pair construction beyond nearest neighbors (e.g., cluster-then-cover, active pairing); (v) evaluate on broader datasets and multilingual settings; and (vi) incorporate human-in-the-loop review and bias audits. These steps aim to strengthen reliability, fairness, and generalizability while keeping the method practical and transparent.

\section{Acknowledgments}
This research was supported in part by the Intramural Research Program of the National Institutes of Health (NIH). The contributions of the NIH authors are considered Works of the United States Government. The findings and conclusions presented in this paper are those of the authors and do not necessarily reflect the views of the NIH or the U.S. Department of Health and Human Services. Jie Shen is supported by NSF-AF-2239376 (CAREER award).
%


\bibliography{fair}
\bibliographystyle{plain}

\newpage
\section*{NeurIPS Paper Checklist}

The checklist is designed to encourage best practices for responsible machine learning research, addressing issues of reproducibility, transparency, research ethics, and societal impact. Do not remove the checklist: {\bf The papers not including the checklist will be desk rejected.} The checklist should follow the references and follow the (optional) supplemental material.  The checklist does NOT count towards the page
limit. 

Please read the checklist guidelines carefully for information on how to answer these questions. For each question in the checklist:
\begin{itemize}
    \item You should answer \answerYes{}, \answerNo{}, or \answerNA{}.
    \item \answerNA{} means either that the question is Not Applicable for that particular paper or the relevant information is Not Available.
    \item Please provide a short (1–2 sentence) justification right after your answer (even for NA). 
\end{itemize}

{\bf The checklist answers are an integral part of your paper submission.} They are visible to the reviewers, area chairs, senior area chairs, and ethics reviewers. You will be asked to also include it (after eventual revisions) with the final version of your paper, and its final version will be published with the paper.

The reviewers of your paper will be asked to use the checklist as one of the factors in their evaluation. While "\answerYes{}" is generally preferable to "\answerNo{}", it is perfectly acceptable to answer "\answerNo{}" provided a proper justification is given (e.g., "error bars are not reported because it would be too computationally expensive" or "we were unable to find the license for the dataset we used"). In general, answering "\answerNo{}" or "\answerNA{}" is not grounds for rejection. While the questions are phrased in a binary way, we acknowledge that the true answer is often more nuanced, so please just use your best judgment and write a justification to elaborate. All supporting evidence can appear either in the main paper or the supplemental material, provided in appendix. If you answer \answerYes{} to a question, in the justification please point to the section(s) where related material for the question can be found.

IMPORTANT, please:
\begin{itemize}
    \item {\bf Delete this instruction block, but keep the section heading ``NeurIPS Paper Checklist"},
    \item  {\bf Keep the checklist subsection headings, questions/answers and guidelines below.}
    \item {\bf Do not modify the questions and only use the provided macros for your answers}.
\end{itemize}


\begin{enumerate}

\item {\bf Claims}
    \item[] Question: Do the main claims made in the abstract and introduction accurately reflect the paper's contributions and scope?
    \item[] Answer:  \answerYes{}
    \item[] Justification: \justificationTODO{}

\item {\bf Limitations}
    \item[] Question: Does the paper discuss the limitations of the work performed by the authors?
    \item[] Answer:  \answerYes{} 
    \item[] Justification: \justificationTODO{}

\item {\bf Theory assumptions and proofs}
    \item[] Question: For each theoretical result, does the paper provide the full set of assumptions and a complete (and correct) proof?
    \item[] Answer:  \answerNA{} 
    \item[] Justification: No theory

    \item {\bf Experimental result reproducibility}
    \item[] Question: Does the paper fully disclose all the information needed to reproduce the main experimental results of the paper to the extent that it affects the main claims and/or conclusions of the paper (regardless of whether the code and data are provided or not)?
    \item[] Answer: \answerYes{} 
    \item[] Justification: \justificationTODO{}

\item {\bf Open access to data and code}
    \item[] Question: Does the paper provide open access to the data and code, with sufficient instructions to faithfully reproduce the main experimental results, as described in supplemental material?
    \item[] Answer: \answerYes{} 
    \item[] Justification: \justificationTODO{}

\item {\bf Experimental setting/details}
    \item[] Question: Does the paper specify all the training and test details (e.g., data splits, hyperparameters, how they were chosen, type of optimizer, etc.) necessary to understand the results?
    \item[] Answer: \answerYes{} 
    \item[] Justification: \justificationTODO{}

\item {\bf Experiment statistical significance}
    \item[] Question: Does the paper report error bars suitably and correctly defined or other appropriate information about the statistical significance of the experiments?
    \item[] Answer: \answerYes{} 
    \item[] Justification: 

\item {\bf Experiments compute resources}
    \item[] Question: For each experiment, does the paper provide sufficient information on the computer resources (type of compute workers, memory, time of execution) needed to reproduce the experiments?
    \item[] Answer: \answerYes{} 
    \item[] Justification: \justificationTODO{}
    
\item {\bf Code of ethics}
    \item[] Question: Does the research conducted in the paper conform, in every respect, with the NeurIPS Code of Ethics \url{https://neurips.cc/public/EthicsGuidelines}?
    \item[] Answer: \answerYes{} 
    \item[] Justification: \justificationTODO{}

\item {\bf Broader impacts}
    \item[] Question: Does the paper discuss both potential positive societal impacts and negative societal impacts of the work performed?
    \item[] Answer: \answerYes{} 
    \item[] Justification: \justificationTODO{}
    
\item {\bf Safeguards}
    \item[] Question: Does the paper describe safeguards that have been put in place for responsible release of data or models that have a high risk for misuse (e.g., pretrained language models, image generators, or scraped datasets)?
    \item[] Answer: \answerYes{} 
    \item[] Justification: \justificationTODO{}

\item {\bf Licenses for existing assets}
    \item[] Question: Are the creators or original owners of assets (e.g., code, data, models), used in the paper, properly credited and are the license and terms of use explicitly mentioned and properly respected?
    \item[] Answer: \answerYes{} 
    \item[] Justification: public dataset

\item {\bf New assets}
    \item[] Question: Are new assets introduced in the paper well documented and is the documentation provided alongside the assets?
    \item[] Answer: \answerYes{} 
    \item[] Justification: will release the code

\item {\bf Crowdsourcing and research with human subjects}
    \item[] Question: For crowdsourcing experiments and research with human subjects, does the paper include the full text of instructions given to participants and screenshots, if applicable, as well as details about compensation (if any)? 
    \item[] Answer: \answerNA{} 
    \item[] Justification: public dataset

\item {\bf Institutional review board (IRB) approvals or equivalent for research with human subjects}
    \item[] Question: Does the paper describe potential risks incurred by study participants, whether such risks were disclosed to the subjects, and whether Institutional Review Board (IRB) approvals (or an equivalent approval/review based on the requirements of your country or institution) were obtained?
    \item[] Answer: \answerNA{} 
    \item[] Justification: public dataset

\item {\bf Declaration of LLM usage}
    \item[] Question: Does the paper describe the usage of LLMs if it is an important, original, or non-standard component of the core methods in this research? Note that if the LLM is used only for writing, editing, or formatting purposes and does not impact the core methodology, scientific rigorousness, or originality of the research, declaration is not required.
    \item[] Answer: \answerNA{} 
    \item[] Justification: only for polishing paper

\end{enumerate}

\end{document}